\documentclass[wcp]{jmlr}

\usepackage{longtable}

\usepackage{booktabs}
\usepackage{times}
\usepackage{soul}
\usepackage{url}

\usepackage[small]{caption}
\usepackage{graphicx}
\usepackage{amsmath}
\usepackage{algorithm}
\usepackage{algorithmic}
\usepackage{xcolor}
\usepackage{adjustbox}
\usepackage{mathtools}
\DeclarePairedDelimiter{\norm}{\lVert}{\rVert}

\makeatletter
\let\Ginclude@graphics\@org@Ginclude@graphics 

\newcommand{\printfnsymbol}[1]{%
  \textsuperscript{\@fnsymbol{#1}}%
}

\usepackage[symbol]{footmisc}

\renewcommand{\thefootnote}{\fnsymbol{footnote}}

\makeatother

\jmlrvolume{189}
\jmlryear{2022}
\jmlrworkshop{ACML 2022}

\title[Towards Data-Free Domain Generalization]{Towards Data-Free Domain Generalization}

  \author{\Name{Ahmed Frikha}\printfnsymbol{1} \Email{ahmed.frikha@siemens.com}\\
  \addr Siemens Technology and University of Munich
  \AND
  \Name{Haokun Chen}\printfnsymbol{1} \Email{haokun.chen@siemens.com}\\
  \addr Siemens Technology and University of Munich
  \AND
  \Name{Denis Krompaß} \Email{denis.krompass@siemens.com}\\
  \addr Siemens Technology
  \AND
  \Name{Thomas Runkler} \Email{thomas.runkler@siemens.com}\\
  \addr Siemens Technology and Technical University of Munich
  \AND
  \Name{Volker Tresp} \Email{volker.tresp@siemens.com}\\
  \addr Siemens Technology and University of Munich
 }

\editors{Emtiyaz Khan and Mehmet G\"{o}nen}

\begin{document}

\maketitle

\begin{abstract}
In this work, we investigate the unexplored intersection of domain generalization (DG) and data-free learning. In particular, we address the question: How can knowledge contained in models trained on different source domains be merged into a single model that generalizes well to unseen target domains, in the absence of source and target domain data? Machine learning models that can cope with domain shift are essential for real-world scenarios with often changing data distributions. Prior DG methods typically rely on using source domain data, making them unsuitable for private decentralized data. We define the novel problem of Data-Free Domain Generalization (DFDG), a practical setting where models trained on the source domains separately are available instead of the original datasets, and investigate how to effectively solve the domain generalization problem in that case. We propose DEKAN, an approach that extracts and fuses domain-specific knowledge from the available teacher models into a student model robust to domain shift. Our empirical evaluation demonstrates the effectiveness of our method which achieves first state-of-the-art results in DFDG by significantly outperforming data-free knowledge distillation and ensemble baselines. \footnotetext[1]{Equal contribution}
\end{abstract}

\section{Introduction}

Deep learning methods have achieved impressive performance in a wide variety of tasks where the data is independent and identically distributed. However, real-world scenarios usually involve a distribution shift between the training data used during development and the test data faced at deployment time. In such situations, deep learning models suffer from a performance degradation and fail to generalize to the out-of-distribution (OOD) data from the target domain \citep{torralba2011unbiased}. For instance, this domain shift problem is encountered on MRI data from different clinical centers that use different scanners \citep{dou2019domain}. Domain Adaptation (DA) approaches \citep{wilson2020survey} assume access to data from the source domain(s) for training as well as target domain data for model adaptation. However, data collection from the target domain can sometimes be expensive, slow, or infeasible, e.g. self-driving cars have to generalize to a variety of weather conditions \citep{zhang2017curriculum} in urban and rural environments from different countries. In this work, we focus on Domain Generalization (DG) \citep{muandet2013domain}, where a model trained on multiple source domains is applied without any modification to unseen target domains. 

In the last decade, a plethora of DG methods requiring only access to the source domains were proposed \citep{zhou2021domain}. Nevertheless, the assumption that access to source domain data can always be granted does not hold in many cases. For instance, General Data Protection Regulation (GDPR) prohibits the access to confidential information and sensitive data that might identify individuals, e.g. bio-metric data. Likewise, some commercial entities are not willing to share their original data to prevent competitive disadvantage. Furthermore, as datasets get larger, their release, transfer, storage and management can become prohibitively expensive \citep{lopes2017data}. To circumvent the concerns related to releasing the original dataset, the data owners might want to share a model trained on their data instead. In light of increasing data privacy concerns, this alternative has recently enjoyed a surge of interest \citep{micaelli2019zero,chen2019data,nayak2019zero,liang2020we,li2020model,kundu2020universal,yin2020dreaming,ahmed2021unsupervised}.

Although Data-Free Knowledge Distillation (DFKD) methods were developed to transfer knowledge from a teacher model to a student model without access to the original data \citep{lopes2017data,micaelli2019zero,chen2019data,nayak2019zero,yin2020dreaming}, only single-teacher scenarios with no domain shift were studied. On the other hand, Source-Free Domain Adaptation (SFDA) approaches were proposed to tackle the domain shift problem setting where one \citep{liang2020we,li2020model,kundu2020universal} or multiple \citep{ahmed2021unsupervised} models trained on source domain data are available instead of the original dataset(s). Nonetheless, they require access to data from the target domain. In this work, we investigate the unstudied intersection of Domain Generalization and Data-Free Learning. Data-Free Domain Generalization (DFDG) is a problem setting that assumes only access to models trained on the source domains, without requiring data from source or target domains. Hereby, the goal is to have a single model able to generalize to unseen domains without any modification or data exposure, as it is the case in DG.

Our contribution is threefold: Firstly, we introduce and define the novel and practical DFDG problem setting. Secondly, we tackle it by proposing a first and strong approach that merges the knowledge stored in the domain-specific models via the generation of synthetic data and distills it into a single model. Thirdly, we demonstrate the effectiveness of our method by empirically evaluating it on two DG benchmark datasets.

\section{Related Work}
\label{related-work}
To the best of our knowledge, we are the first to address the Data-Free Domain Generalization (DFDG) problem. In the following, we discuss approaches to related problem settings.

\subsection{Domain Generalization}
Domain Generalization (DG) approaches can be classified into three categories. Domain alignment methods attempt to learn a domain-invariant representation of the data from the source domains by regularizing the learning objective. Variants of such a regularization include the minimization across the source domains of the maximum mean discrepancy criteria (MMD) \citep{gretton2012kernel,li2018domain}, the minimization of a distance metric between the domain-specific means \citep{tzeng2014deep} or covariance matrices \citep{sun2016deep}, the minimization of a contrastive loss \citep{motiian2017unified,kim2021selfreg}, or the maximization of loss gradient alignment \citep{shi2021gradient,shahtalebi2021sand}. Other works use adversarial training with a domain discriminator model \citep{ganin2016domain,li2018deep} for the same purpose. Another category of works leverages meta-learning techniques, e.g., the bi-level optimization scheme proposed in \citep{finn2017model}, to optimize for adaptation. Even though gradient-based meta-learning methods \citep{finn2017model} were initially developed to enable few-shot learning, they were leveraged to address a wide variety of problems, such as continual learning \citep{riemer2018learning, frikha2021arcade}, anomaly detection \citep{frikha2021few} and DG. In DG, meta-learning was employed to optimize for quick adaptation to different domains \citep{li2018learning}, to learn how to regularize the output layer \citep{balaji2018metareg}, and to regularize the embedding space \citep{dou2019domain}. Another line of work augment the training data to tackle DG. Hereby, the source domain data is perturbed by computing inter-domain examples \citep{xu2020adversarial,yan2020improve,wang2020heterogeneous} via Mixup \citep{zhang2017mixup}, by randomizing the style of images \citep{nam2019reducing}, by computing adversarial examples \citep{goodfellow2014explaining} using a class classifier \citep{sinha2017certifying,qiao2020learning} or a domain classifier \citep{shankar2018generalizing}, or corrupting learned features to incentivize new feature discovery \citep{frikha2021columbus}.
Other works perturb intermediate representations of the data \citep{huang2020self,zhou2021mixstyle,frikha2021columbus}. Unlike standard DG approaches that require access to the source domain datasets, our method merges the domain-specific knowledge from models trained on the source domains into a single model resilient to domain shift, while preserving data privacy (Figure \ref{fig:dfdg}).
\subsection{Knowledge Distillation}

Knowledge distillation (KD) \citep{hinton2015distilling} was originally proposed to compress the knowledge of a large teacher network into a smaller student network. Several KD extensions and improvements enabled its application to a variety of scenarios including quantization \citep{mishra2017apprentice}, domain adaptation \citep{zhao2020multi}, and few-shot learning \citep{rajasegaran2020self}. While these methods rely on the original data, Data-Free Knowledge Distillation (DFKD) methods were recently developed \citep{lopes2017data,micaelli2019zero,nayak2019zero,chen2019data, liu2021data}. Hereby, knowledge is transferred from one \citep{micaelli2019zero,nayak2019zero,chen2019data,choi2020data,yin2020dreaming,luo2020large,zhang2021diversifying} or multiple \citep{li2021mixmix} teacher(s) to the student model via the generation of synthetic data, either by optimizing random noise examples \citep{nayak2019zero,yin2020dreaming,zhang2021diversifying} or by training a generator \citep{micaelli2019zero,chen2019data,choi2020data,luo2020large}. Nevertheless, the aforementioned DFKD methods focus on scenarios without any domain shift, i.e. the student is evaluated on examples from the same data distribution used for training the teacher. In the DFDG problem setting we address, the student is trained from multiple teachers that are trained on different source domains in a way that enables generalization to data from unseen target domains. We propose a baseline that extends the usage of a recent DFKD method \citep{zhang2021diversifying} to the DFDG setting, and compare it to our approach (Section \ref{results}).

\subsection{Source-free domain adaptation}

The recently addressed Source-Free Domain Adaptation problem \citep{liang2020we,li2020model,kundu2020universal} assumes access to one or multiple model(s) trained on the source domains, as well as data examples from a specific target domain. Proposed approaches to tackle it include the combination of generative models with a regularization loss \citep{li2020model}, a feature alignment mechanism \citep{yeh2021sofa}, or a weighting of the target domain samples by their similarity to the source domain \citep{kundu2020universal}. SHOT \citep{liang2020we} employs an information maximization loss along with a self-supervised pseudo-labeling, and is extended to the multi-source scenario via source model weighting \citep{ahmed2021unsupervised}. BUFR \citep{eastwood2021source} aligns the target domain feature distribution with the one from the source domain. Another line of works leverage Batch Normalization (BN) \citep{ioffe2015batch} layers by replacing the BN-statistics computed on the source domain with those computed on the target domain \citep{li2016revisiting}, or by training the BN-parameters on the target domain via entropy minimization \citep{wang2020tent}. While these approaches rely on the availability of data from a known target domain, we address the DFDG scenario where the model is expected to generalize to \emph{a priori unknown} target domain(s) without any modification or exposure to their data. We also note that some methods \citep{kundu2020universal,liang2020we,eastwood2021source} modify the training procedure on the source domain, which would not be possible in cases where the data is not accessible anymore. 
\subsection{Federated Learning}
Federated Learning (FL) \citep{mcmahan2017communication} is a decentralized learning paradigm where multiple clients train deep learning models without centralizing their local data. While most FL works \citep{yin2021comprehensive} focus on single-domain applications, i.e., the clients have the same data distribution, concurrent works addressed FL settings involving distribution shift \citep{zhang2021federated, liu2021feddg, li2021fedbn, chen2022fraug}. We propose DFDG as an alternative privacy-preserving DG setting, which enables cross-domain knowledge sharing in a static, permanent and communication-efficient way, i.e., no client interaction overhead. Moreover, in scenarios where the data cannot be accessed, e.g., it was lost or deleted, or only a trained model was made public, FL is not applicable as it requires access to the data. However, DFDG methods would be applicable.

\section{Approach}

\subsection{Problem statement}

Let $D_{s}$ and $D_{t}^{j}$ denote the datasets available from the source and target domains respectively with $i=1, .., I$ and $j=1, .., J$. Hereby, $I$ and $J$ denote the number of source and target domains respectively. In Domain Generalization (DG), the goal is to train a model on the source domain data $D_{s}^{i}$ in a way that enables generalization to \textit{a priori} unavailable target domain data $D_{t}^{j}$, without any model modification at test time. We consider the source-data-free scenario of this problem where the source domain datasets $D_{s}^{i}$ are not accessible, e.g., due to privacy, security, safety or commercial concerns, and models trained on these domain-specific datasets separately are available instead. 

\begin{figure}[ht]
  \centering
  \includegraphics[width=\textwidth]{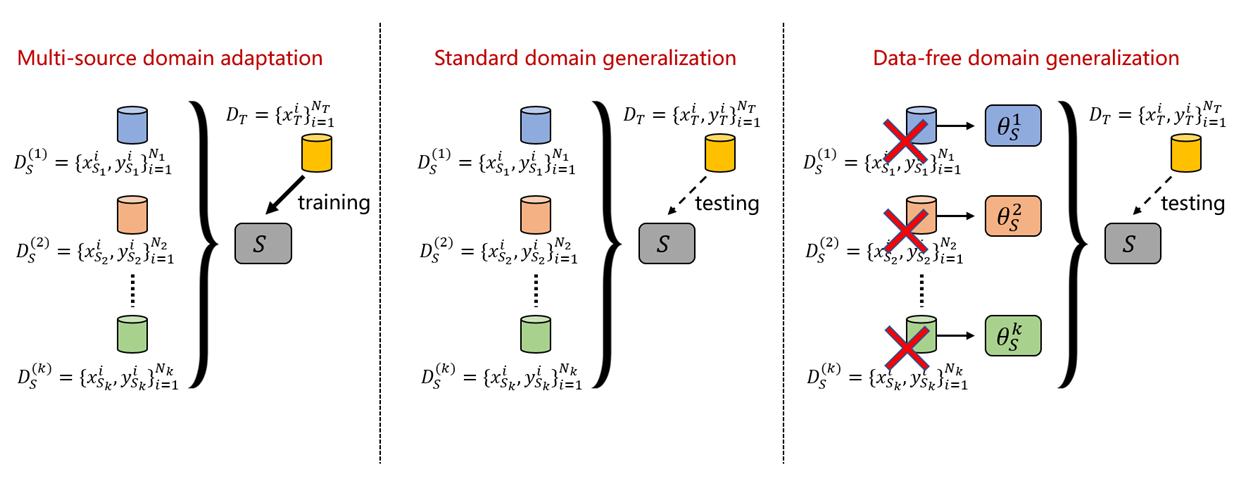}
  \caption{Comparison between the domain generalization, the multi-source domain adaptation and the proposed data-free domain generalization problems.}
  \label{fig:dfdg}
\end{figure}

We refer to the source domain models as teacher models $T_{i}$ as in the knowledge distillation literature \citep{hinton2015distilling}. We assume that the teacher models were trained without the prior knowledge that they would be used in a DFDG setting, i.e., their training does not involve any domain shift robustness mechanism. Hence, the application scenarios where the source domain data is not accessible anymore, e.g., was deleted, are also considered. We refer to this novel learning scenario as \emph{Data-Free Domain Generalization (DFDG)}. The major difference with Source-Free Domain Adaptation (SFDA) \citep{liang2020we,li2020model,kundu2020universal} is the absence of target domain data $D_{t}^{j}$ available for training in DFDG (Figure \ref{fig:dfdg}). 

The DFDG problem is a prototype for a practical use case where a model robust to domain shift is needed and models trained on the same task but different domains are available. This problem definition is motivated by the question: How can we merge the knowledge from multiple models trained on different domains into a single model that is able to generalize to unseen target domains without any data exposure?

Applications of the proposed DFDG include every DG application that involves inaccessible data, e.g., due to privacy concerns, every DFKD application (see Table 4 in \cite{liu2021data} for an overview) that involves domain shift, and every SFDA application with data-scarce or \emph{a priori} unknown target domains. DFDG addresses all real-world scenarios where different entities have the same application with different data distributions, are unwilling (or not allowed) to share their original data, and would benefit from a model robust to domain shift. In the healthcare sector, the entities could be clinical centers (CCs) using different scanners or data acquisition protocols, such as in \citep{dou2019domain}. In the latter, the MRI datasets are not made public, probably due to patient privacy concerns. With DFDG promoting a privacy-preserving way of data sharing, the CCs might consider releasing models. Analogously, in industrial manufacturing, companies are usually unwilling to share their raw production data, due to concerns of intellectual property infringement and reverse engineering. In this sector, examples of DFDG applications include image classification of gas turbine deficiencies and powder bed anomalies in additive manufacturing. Hereby, the images are taken in diverse production environments and conditions specific to the data owner. Note that a model robust to domain shift would be desirable for the machine users, i.e., data owners, and/or for the machine manufacturer (of the gas turbine or additive manufacturing machine).

\subsection{DEKAN: Domain Entanglement via Knowledge Amalgamation from Domain-Specific Networks} \label{dekan}

To address the DFDG problem, we propose Domain Entanglement via Knowledge Amalgamation from domain-specific Networks (DEKAN). Our approach tackles the different challenges of DFDG in 3 stages: Knowledge extraction, fusion and transfer. In the first stage, we extract the knowledge from the different source domain teacher models separately by generating domain-specific synthetic datasets. Thereafter, DEKAN generates cross-domain synthetic data by leveraging all pairs of inter-domain model-dataset combinations. Hereby, the cross-domain examples are optimized to be recognizable by teacher models trained on different domains. In the final stage, DEKAN transfers the extracted knowledge from the domain-specific teachers to a student model via knowledge distillation using the generated data. At test time, the resulting student model is evaluated on target domain data without any modification. In the following, we introduce the stages in more detail. 

\subsubsection{Intra-Domain Data-Free Knowledge Extraction}

In this stage, we extract the domain-specific knowledge from the available teacher models $T_{i}$ separately by generating domain-specific synthetic datasets $D_{g}^{i}$. For this, we apply an improved version \citep{zhang2021diversifying} of the data-free knowledge distillation method DeepInversion (DI) \citep{yin2020dreaming} that enables the generation of more diverse images. Hereby, we use inceptionism-style \citep{mahendran2015understanding} image synthesis, also called DeepDream or inversion, i.e., we initialize random noise images $\hat{x}$ and optimize them to be recognized as samples from pre-defined classes by a trained model. We uniformly sample labels $y$ and optimize the corresponding random images $\hat{x}$ by minimizing the domain-specific inversion loss $L_{DS}$ given by

\begin{equation} \label{eq:imggen1}
L_{DS} = L_{C}(T(\hat{x}), y) + \lambda_{1}L_{R}(\hat{x}) + \lambda_{2}L_{M}(\hat{x}),
\end{equation}

\noindent where $L_{C}$ denotes the classification loss, e.g., cross-entropy, $L_{R}$ an image prior regularization, $L_{M}$ a feature moment matching loss, and $\lambda_{1}$ and $\lambda_{2}$ weighting coefficients. $L_{R}$ penalizes the $l_2$-norm and the total variation of the image to ensure the convergence to valid natural images \citep{mahendran2015understanding,yin2020dreaming}. $L_{M}$, also called moment matching loss, optimizes the synthetic images so that their feature distributions captured by batch normalization (BN) layers match those of the real data used to train the teacher model. Formally, 

\begin{equation} \label{eq:mm}
    L_{M}(\hat{x})= \sum_{l} max(\norm{\mu_l(\hat{x}) - \hat{\mu_l}}_{2} - \delta_l, 0)+ \sum_{l} max(\norm{\sigma_l^2(\hat{x}) - \hat{\sigma_l}^2}_{2} - \gamma_l, 0).
\end{equation}

$L_{M}$ minimizes the $l_2$-norm between the BN-statistics of the synthetic data, i.e., mean $\mu_l(\hat{x})$ and variance $\sigma_l^2(\hat{x})$, and those stored in the trained teacher model, $\hat{\mu_l}$ and $\hat{\sigma_l}^2$, at each BN layer $l$ \citep{yin2020dreaming}. In order to increase the diversity of the generated images, we relax this optimization by allowing the BN-statistics computed on the synthetic images to deviate from those stored in the model within certain margins, as introduced in \citep{zhang2021diversifying}. These deviation margins are defined by relaxation constants for mean and variance, denoted by $\delta_l$ and $\gamma_l$ respectively. The latter are computed as the $\epsilon_{DS}$ percentile of the distribution of differences between the stored BN-statistics and those computed using random images, as proposed in \citep{zhang2021diversifying}. We note that the higher the value of the hyperparameter $\epsilon_{DS}$, the higher the relaxation. 

We apply this data-free inversion to each model $T_{i}$ separately, yielding datasets $D_{g}^{i}$ that are correctly classified by the respective model and match the distribution of features extracted by it.

\subsubsection{Cross-Domain Data-Free Knowledge Fusion}

\begin{figure}[th]
  \centering
  \includegraphics[width=\textwidth]{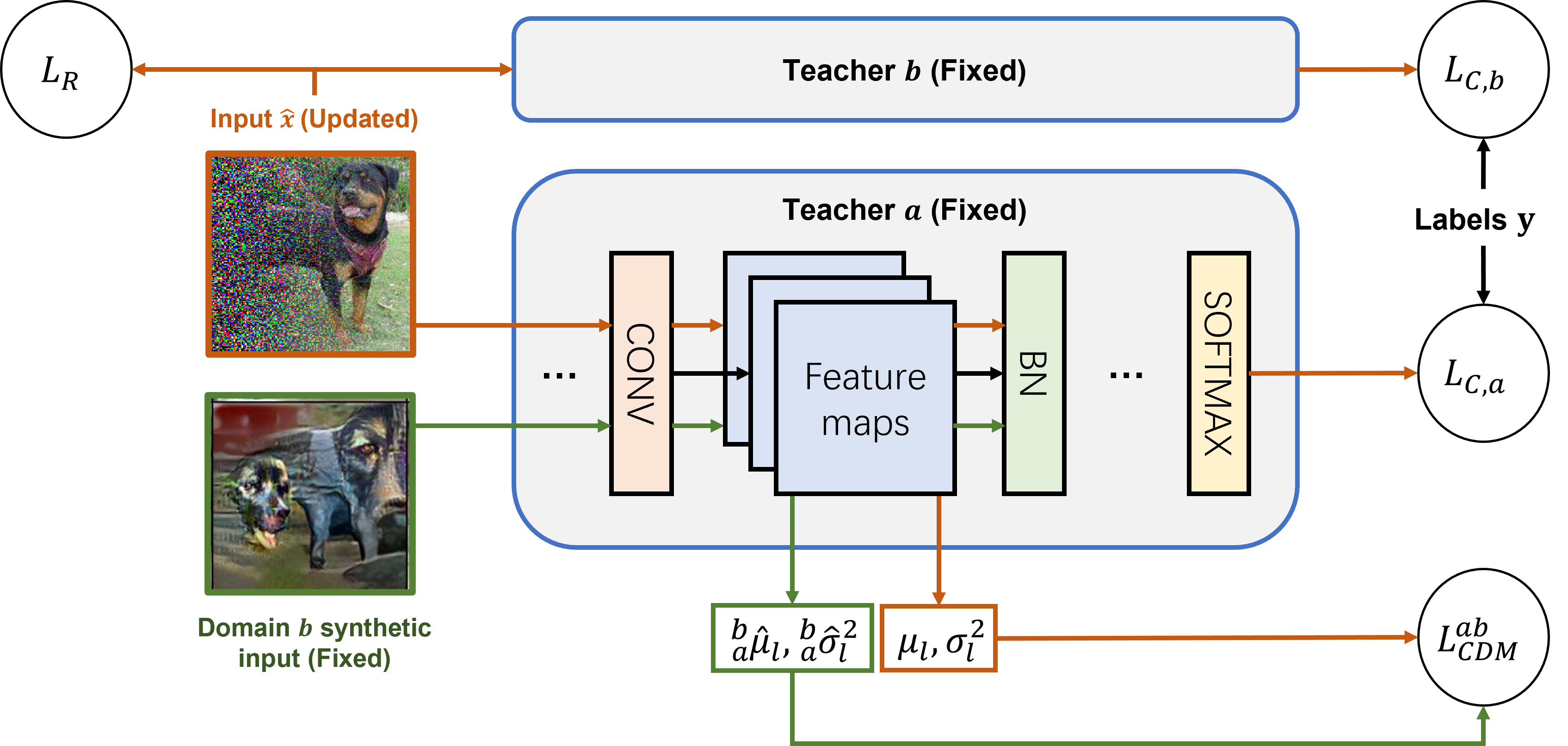}
  \caption{Overview of the Cross-Domain Data-Free Knowledge Fusion.}
  \label{fig:cd}
\end{figure}

In the second stage, we propose a technique to merge the knowledge from two domains by generating cross-domain synthetic images that capture class-discriminative features present in the two domains, and match the distribution of intermediate features extracted by a domain-specific model from images of another domain. Let $T_{a}$ and $T_{b}$ denote the teacher models, and $D_{g}^{a}$ and $D_{g}^{b}$ the synthetic data generated in the first stage, specific to two domains $a$ and $b$. We generate synthetic images $D_{g}^{ab}$ by minimizing the cross-domain inversion loss $L_{CD}^{ab}$, that we formulate as 

\begin{equation} \label{eq:imggen2}
    L_{CD}^{ab} = L_{C}(T_{a}(\hat{x}), y) + L_{C}(T_{b}(\hat{x}), y) + \alpha_{1}L_{R}(\hat{x}) + \alpha_{2}L_{CDM}^{ab}(\hat{x}),
\end{equation}

\noindent where $L_{C}$ denotes the classification loss, e.g., cross-entropy, $L_{R}$ the aforementioned image prior regularization, $L_{CDM}$ the cross-domain feature moment matching loss, and $\alpha_{1}$ and $\alpha_{2}$ weighting coefficients. We incentivize the generated images to contain class-discriminative features from both domains by minimizing the classification loss using both teachers. We hypothesize that images that can be recognized by models trained on different domains capture more domain-agnostic semantic features than those generated by inverting a single domain-specific model as done in prior works. In addition, the cross-domain feature distribution matching loss $L_{CDM}^{ab}$ optimizes the cross-domain synthetic images $D_{g}^{ab}$ so that their feature distribution matches the distribution of the features extracted by $T_a$, the model trained on domain $a$, for images $D_{g}^{b}$ synthesized from domain $b$. Note that $L_{CDM}^{ab} \neq L_{CDM}^{ba}$ and that using the model $T_b$ and the data generated by inverting $T_a$ in the first stage, i.e., $D_{g}^{a}$, would yield the cross-domain images $D_{g}^{ba}$ that are different from $D_{g}^{ab}$. Formally, 


\begin{equation} \label{eq:cdm}
    L_{CDM}^{ab}(\hat{x})= \sum_{l} max(\norm{\mu_l(\hat{x}) - {}_{a}^{b}\hat{\mu}_{l}}_{2} - {}_{a}^{b}\delta_{l}, 0) + \sum_{l} max(\norm{\sigma_l^2(\hat{x}) - {}_{a}^{b}\hat{\sigma}_{l}^2}_{2} - {}_{a}^{b}\gamma_{l}, 0).
\end{equation}

Similarly to $L_{M}$ (Eq. \ref{eq:mm}) in the first stage, $L_{CDM}^{ab}$ minimizes the $l_2$-norm between the BN-statistics of the synthetic data, $\mu_l(\hat{x})$ and $\sigma_l^2(\hat{x})$, and target statistics, at each BN layer $l$. Here, the target statistics, ${}_{a}^{b}\hat{\mu}_{l}$ and ${}_{a}^{b}\hat{\sigma}_{l}^2$, are computed in a way that involves knowledge from different domains. In particular, they result from feeding the synthetic data specific to domain $b$ to the teacher trained on data from domain $a$, and computing the first two feature moments, i.e., mean and variance, for each BN layer. The intention behind this is to synthesize images that capture the features learned by the model on domain $a$ that are activated and recognized when exposed to images from domain $b$. We hypothesize that such images would encompass domain-agnostic semantic information that would be useful for training a single model resilient to domain shift in the next stage.

We relax $L_{CDM}$ by allowing the BN-statistics of the synthetic input to fluctuate within a certain interval. Here, we compute the relaxation constants ${}_{a}^{b}\delta_{l}$ and ${}_{a}^{b}\gamma_{l}$ as the $\epsilon_{CD}$ percentile of the distribution of differences between the stored BN-statistics, i.e., computed on the original domain $a$ images, and those computed using the images $D_{g}^{b}$ synthesized from the domain $b$ teacher in the first stage. $\epsilon_{CD}=100\%$ corresponds to synthesized images $\hat{x}$ yielding the BN-statistics from domain $a$, i.e., stored in model $T_a$, would not be penalized, i.e., $L_{CDM}^{ab}=0$. This stage can be viewed as a domain augmentation, since the synthesized images $D_{g}^{ab}$ do not belong neither to domain $a$ nor to domain $b$. The synthesis of cross-domain data is applied to all possible domain pairs.

\begin{algorithm}
\caption{Domain Entanglement via Knowledge Amalgamation from domain-specific Networks}
\begin{algorithmic}[1]
\REQUIRE $T_{1 .. I}$: $I$ Domain-specific teacher models\\
  // First stage: Intra-Domain Knowledge Extraction
  \FOR{$i \gets 1$ to $I$}
    \STATE Initialize the domain-specific synthetic data $D_{g}^{i}$: Images $\hat{x}\sim\mathcal{N}$(0, $I$) and arbitrary labels
    \WHILE{not converged}
      \STATE Update $D_{g}^{i}$ by minimizing the domains-specific inversion loss $L_{DS}$ (Eq. \ref{eq:imggen1}) using $T_{i}$ 
    \ENDWHILE \\
  \ENDFOR\\
  // Second stage: Cross-Domain Knowledge Fusion
  \FOR{$i \gets 1$ to $I$}
    \FOR{$j \gets 1$ to $I$ and $i \neq j$}
      \STATE Initialize the cross-domain synthetic data $D_{g}^{ij}$: Images $\hat{x}\sim\mathcal{N}$(0, $I$) and arbitrary labels
      \WHILE{not converged}
        \STATE Update $D_{g}^{ij}$ by minimizing the cross-domain loss $L_{CD}^{ij}$ (Eq. \ref{eq:imggen2}) using $T_{i}$, $T_{j}$ and $D_{g}^{j}$
     \ENDWHILE \\
    \ENDFOR
  \ENDFOR\\
  // Third stage: Multi-Domain Knowledge Distillation
  \STATE Initialize the student model $S_{\theta}$ randomly or from a pre-trained model
  \STATE Concatenate the domain-specific and cross-domain synthetic datasets into one dataset $D_{g}$
  \WHILE{not converged}
    \STATE Randomly sample a mini-batch $B=\{\hat{x}, y\}$ from $D_{g}$
    \STATE Update $\theta$ by minimizing the knowledge distillation loss $L_{KD}$ (Eq. \ref{eq:kd_loss}) using $B$ and $T_{1 .. I}$
  \ENDWHILE\\
\STATE \textbf{return} Domain-generalized student model $S_{\theta}$
\end{algorithmic}
\end{algorithm}

\subsubsection{Multi-Domain Knowledge Distillation}

In the final DEKAN stage, the domain-specific and cross-domain knowledge, which is captured in the synthetic data generated in the first and second stages respectively, is transferred to a single student model $S$. To this end, we use knowledge distillation \citep{hinton2015distilling}, i.e., we train the student model to mimic the predictions of the teachers for the synthetic data.

\begin{equation} \label{eq:kd_loss}
L_{KD} = D_{KL}(S(\hat{x})\;||\; p) \;
\textrm{with} \; p= 
\begin{cases}
    T_{i}(\hat{x}),& \textrm{if } \hat{x} \in D_{g}^{i} \; \textrm{(domain-specific)}\\
    \frac{1}{2}(T_{i}(\hat{x})+T_{j}(\hat{x})),& \textrm{if } \hat{x} \in D_{g}^{ij} \; \textrm{(cross-domain)}
\end{cases}
\end{equation}

As described in Equation \ref{eq:kd_loss}, we minimize the Kullback-Leibler divergence $D_{KL}$ between the predictions of the student $S$ and the teacher(s) corresponding to the synthetic images $\hat{x}$. In particular, if the data examples are domain-specific, i.e., they were generated in the first DEKAN stage, the predictions of the corresponding teacher are used as soft labels to train the student. For the cross-domain synthetic images that were generated in the second stage, the average predictions of the two corresponding teachers is used instead. The aggregation of the prediction distributions of two domain-specific teacher models contributes to the knowledge amalgamation across domains.

Algorithm 1 summarizes the 3 stages of the DEKAN's training procedure. We note that the updates of the syntehtic data and the student model parameters $\theta$ are performed using gradient-based optimization, specifically Adam \citep{kingma2014adam} in our case. Explicit update rule formulas and iteration over the synthetic data batches are omitted for simplicity of notation.

\section{Experiments and Results} \label{results}

The conducted experiments \footnote{Code available under: \url{https://github.com/HaokunChen245/DFDG}} aim to tackle the following questions: $(a)$ How does DEKAN compare to leveraging the domain specific models directly to make predictions on data from unseen domains? $(b)$ How does our approach compare to data-free knowledge distillation methods applied to each domain separately? $(c)$ How much does the unavailability of data cost in terms of performance?

We design baseline methods to address the novel DFDG problem, and compare them with DEKAN. The first category of baselines applies the available domain-specific models on the data from the target domains (Question $(a)$). We consider two ensemble baselines that aggregate the predictions of these models, e.g., by taking the average of the model predictions (\textbf{AvgPred}), or by taking the prediction of the most confident model, i.e., the model with the lowest entropy (\textbf{HighestConf}). Besides, we implement oracle methods that evaluate each of the domain-specific models separately on the target domain and then report the results of the best model (\textbf{BestTeacher}). Furthermore, we propose a baseline that applies an improved version \citep{zhang2021diversifying} of DeepInversion (DI) \citep{yin2020dreaming} on each of the models separately to generate domain-specific synthetic images used to then train a student model via knowledge distillation (\textbf{Multi-DI}; Question $(b)$). Note that Multi-DI is equivalent to the application of DEKAN's first and third stage. Finally, we compare DEKAN to an upper-bound baseline that uses the original data from the source domains to train a single model via Empirical Risk Minimization (\textbf{ERM}) \citep{gulrajani2020search}, a common domain generalization baseline (Question $(c)$).

\begin{table}[h]
\centering
\small
\resizebox{0.7\linewidth}{!}{
\begin{tabular}{lccccc}
\toprule
\textbf{Algorithm}   & \textbf{Art}           & \textbf{Cartoon}           & \textbf{Photo}           & \textbf{Sketch}           & \textbf{Avg}         \\
\midrule
Ens-AvgPred & 79.9 & 65.4 & 96.4 & 79.5 & 80.3 \\
Ens-HighestConf & 82.3 & 66.0 & 96.6 & 76.9 & 80.4 \\
Multi-DI & 82.1 & 72.1 & 95.6 & 73.8 & 80.9 \\
DEKAN (ours) & 83.0 & 75.9 & 96.3 & 80.2 & \textbf{83.9}  \\
\midrule
BestTeacher (oracle) & 75.2 & 62.8 & 96.4 & 69.8 & 76.1  \\
ERM (not data-free) & 86.0 & 81.8 & 96.8 & 80.4 & 86.2\\
\bottomrule
 & &  &  &  & \\
\end{tabular}
}
\resizebox{0.7\linewidth}{!}{
\begin{tabular}{lccccc}
\toprule
\textbf{Algorithm}   & \textbf{MNIST}           & \textbf{M-M}           & \textbf{SVHN}           & \textbf{USPS}           & \textbf{Avg}         \\
\midrule
Ens-AvgPred & 97.9 & 45.8 & 31.3 & 96.1 & 67.8 \\
Ens-HighestConf & 98.5 & 46.7 & 30.5 & 96.5 & 68.0 \\
Multi-DI & 93.3 & 54.0 & 36.7 & 96.5 & 70.2\\
DEKAN (ours) & 94.6 & 55.9 & 39.2 & 96.8 & \textbf{71.6}  \\
\midrule
BestTeacher (oracle) & 99.3 & 48.3 & 38.1 & 97.7 & 70.9 \\
ERM (not data-free) & 98.2 & 55.2 & 50.1 & 96.5 & 75.0   \\
\bottomrule
\end{tabular}
}
\caption{Domain Generalization results on PACS (top) and Digits (bottom).}
\label{r_dg}
\end{table}

We evaluate DEKAN and the baselines on two DG benchmark datasets. PACS  \citep{li2017deeper} includes images that belong to 7 classes from the domains Art-Painting, Cartoon, Photo and Sketch. Digits comprises images four digits datasets: MNIST \citep{lecun1998gradient}, MNIST-M \citep{ganin2015unsupervised}, SVHN \citep{netzer2011reading} and USPS \citep{hull1994database}. We use the same model architecture for the teacher and the student models: ResNet-50 \citep{he2016deep} and ResNet-18 pretrained on ImageNet \citep{russakovsky2015imagenet}, for PACS and Digits respectively. In both synthetic input generation stages, we augment the optimized images before feeding them to the teacher(s). In particular, we apply random horizontal flipping and jitter as done in \citep{yin2020dreaming}, as well as cutout \citep{devries2017improved}, which was found to be essential for data-free object detection \citep{chawla2021data}. For the Digits dataset, random horizontal flipping was not used, as it leads to images of invalid digits. Table \ref{r_dg} shows the results of DEKAN and the baselines. Hereby, the column name refers to the unseen target domain, i.e., the 3 other domains are the source domains used to train the teacher models. The test accuracy is computed on the test set of the target domain.

DEKAN outperforms all data-free baselines on both datasets on average, setting a first state-of-the-art performance for the novel DFDG problem. We find that generative approaches, i.e., Multi-DI and DEKAN, outperform the ensemble methods on average, suggesting that training a single model on data from different domains enables a better aggregation of knowledge than the aggregation of domain-specific model predictions. Most importantly, DEKAN substantially outperforms Multi-DI, highlighting the importance of the synthesized cross-domain images. This is especially the case for the challenging domains, i.e., the domains where all the methods yield the lowest performance. In particular, the generation of cross-domain synthetic data leads to performance improvements of $6.4\%$ and $3.8\%$ on the Sketch and Cartoon PACS domains respectively, as well as a $2.4\%$ increase on the SVHN domain of Digits. Additionally, we note the positive knowledge transfer across domains on the PACS dataset, as all the multi-domain methods outperform the oracle BestTeacher baseline that uses a single domain-specific teacher model, i.e., the teacher that achieves the highest performance on a validation set from the target domain. Finally, it is worth noting that while DEKAN significantly reduces the gap between the best data-free baseline and the upper-bound baseline that uses the original data, there is still potential for improvement.

\begin{figure}[th]
  \centering
  \includegraphics[width=\textwidth]{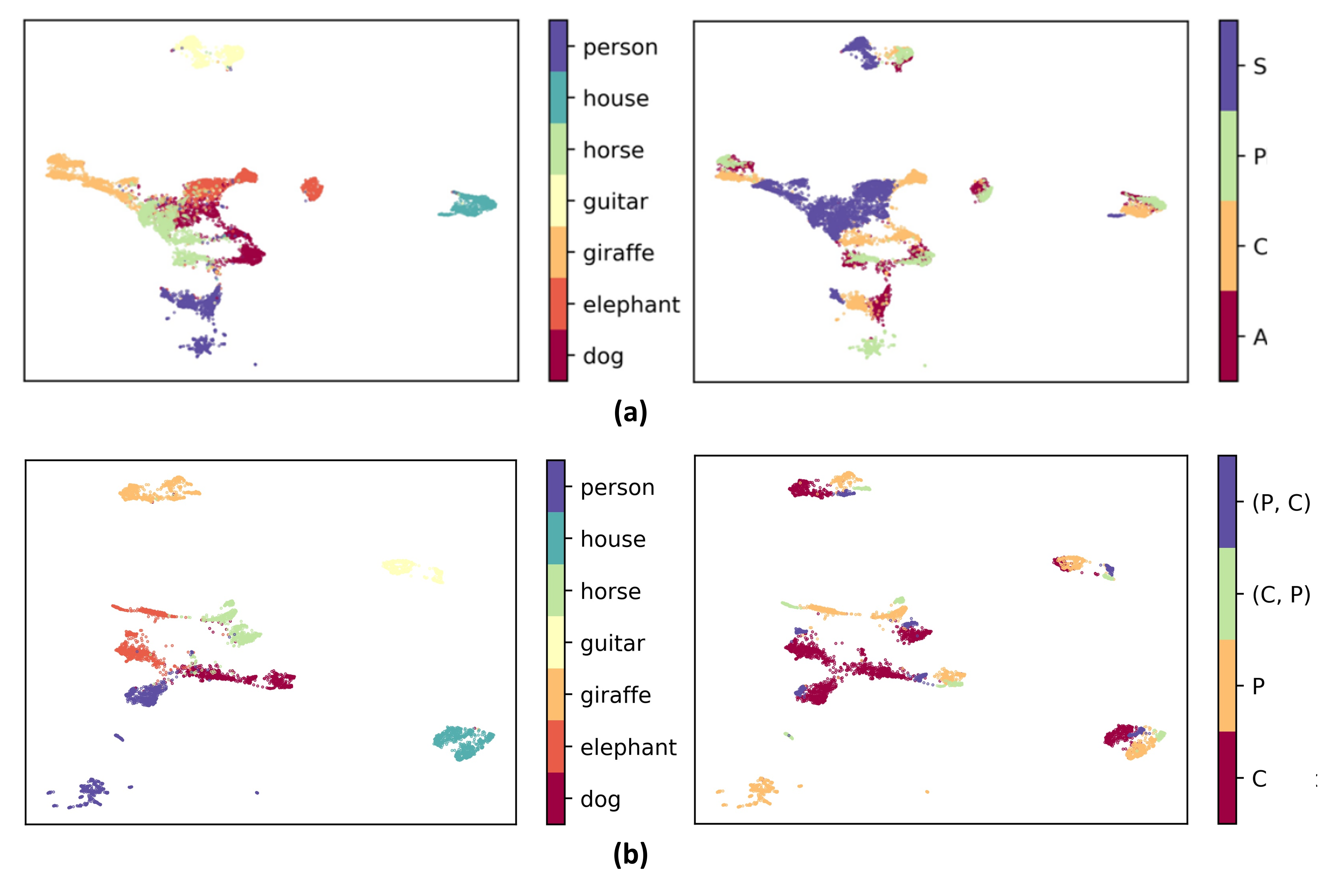}
  \caption{UMAP visualizations of the representations extracted by the last ResNet block, i.e., the input of the linear classifier, of student models trained by DEKAN. Different colors indicate different classes or different domains (S=Sketch, P=Photo, C=Cartoon, A=Art painting). (a) Embeddings of the \emph{original} PACS dataset extracted by the DEKAN-trained student in the setting where the target domain is Sketch. Note that the student is trained exclusively with \emph{synthetic} examples. (b) Embeddings of the original examples from the Photo and Cartoon domains as well as the generated cross-domain images extracted by the DEKAN-trained student in the setting where the target domain is Art painting. (P, C) and (C, P) denote the generated cross-domain synthetic datasets $D_{g}^{PC}$ and $D_{g}^{CP}$, respectively.}
  \label{fig:umap}
\end{figure}


We use UMAP \citep{mcinnes2018umap} to analyze the embeddings extracted by the student model trained by DEKAN (Figure \ref{fig:umap}). Despite being trained exclusively on synthetic data, the student model achieves a good class separation of the embeddings extracted for the original PACS dataset (Figure \ref{fig:umap} (a)). This indicates that the generated data captures class-discriminative features useful for the classification of original examples. We also observe a good separability for most classes of the target domain (Sketch), which is unseen during training. This emphasizes the high domain generalization ability of the student trained with DEKAN, without any access to original data. 

Finally, we investigate the embeddings extracted for the generated cross-domain examples by the student model (Figure \ref{fig:umap} (b)). We find that the embeddings extracted for the generated cross-domain examples lie between the embeddings of the original domain-specific examples, for most classes, e.g., house, person and dog. We hypothesize that the substantial increase in DG performance between DEKAN and Multi-DI (Table \ref{r_dg}) is due to the fact that the generated cross-domain images bridge the gap between the source domains in the student's embedding space.

\section{Conclusion}
This work addresses the unstudied intersection of domain generalization and data-free learning, a practical setting where a model robust to domain shift is needed and the available models were trained on the same task but with data from different domains. We proposed DEKAN, an approach that fuses domain-specific knowledge from the available teacher models into a single student model that can generalize to data from \textit{a priori} unknown domains. Our empirical evaluation demonstrated the effectiveness of our method which outperformed ensemble and data-free knowledge distillation baselines, hence achieving first state-of-the-art results in the novel and challenging data-free domain generalization problem. An interesting avenue for future works could be the integration of differential privacy mechanism into DEKAN. 


\bibliography{acml22}

\begin{thebibliography}{76}
\providecommand{\natexlab}[1]{#1}
\providecommand{\url}[1]{\texttt{#1}}
\expandafter\ifx\csname urlstyle\endcsname\relax
  \providecommand{\doi}[1]{doi: #1}\else
  \providecommand{\doi}{doi: \begingroup \urlstyle{rm}\Url}\fi

\bibitem[Ahmed et~al.(2021)Ahmed, Raychaudhuri, Paul, Oymak, and
  Roy-Chowdhury]{ahmed2021unsupervised}
Sk~Miraj Ahmed, Dripta~S Raychaudhuri, Sujoy Paul, Samet Oymak, and Amit~K
  Roy-Chowdhury.
\newblock Unsupervised multi-source domain adaptation without access to source
  data.
\newblock In \emph{CVPR}, 2021.

\bibitem[Balaji et~al.(2018)Balaji, Sankaranarayanan, and
  Chellappa]{balaji2018metareg}
Yogesh Balaji, Swami Sankaranarayanan, and Rama Chellappa.
\newblock Towards domain generalization using meta-regularization.
\newblock \emph{NeurIPS}, 2018.

\bibitem[Chawla et~al.(2021)Chawla, Yin, Molchanov, and
  Alvarez]{chawla2021data}
Akshay Chawla, Hongxu Yin, Pavlo Molchanov, and Jose Alvarez.
\newblock Data-free knowledge distillation for object detection.
\newblock In \emph{IEEE/CVF Winter Conference on Applications of Computer
  Vision}, 2021.

\bibitem[Chen et~al.(2019)Chen, Wang, Xu, Yang, Liu, Shi, Xu, Xu, and
  Tian]{chen2019data}
Hanting Chen, Yunhe Wang, Chang Xu, Zhaohui Yang, Chuanjian Liu, Boxin Shi,
  Chunjing Xu, Chao Xu, and Qi~Tian.
\newblock Data-free learning of student networks.
\newblock In \emph{ICCV}, 2019.

\bibitem[Chen et~al.(2022)Chen, Frikha, Krompass, and Tresp]{chen2022fraug}
Haokun Chen, Ahmed Frikha, Denis Krompass, and Volker Tresp.
\newblock Fraug: Tackling federated learning with non-iid features via
  representation augmentation.
\newblock \emph{arXiv preprint arXiv:2205.14900}, 2022.

\bibitem[Choi et~al.(2020)Choi, Choi, El-Khamy, and Lee]{choi2020data}
Yoojin Choi, Jihwan Choi, Mostafa El-Khamy, and Jungwon Lee.
\newblock Data-free network quantization with adversarial knowledge
  distillation.
\newblock In \emph{CVPR Workshops}, 2020.

\bibitem[DeVries and Taylor(2017)]{devries2017improved}
Terrance DeVries and Graham~W Taylor.
\newblock Improved regularization of convolutional neural networks with cutout.
\newblock \emph{Arxiv}, 2017.

\bibitem[Dou et~al.(2019)Dou, Coelho~de Castro, Kamnitsas, and
  Glocker]{dou2019domain}
Qi~Dou, Daniel Coelho~de Castro, Konstantinos Kamnitsas, and Ben Glocker.
\newblock Domain generalization via model-agnostic learning of semantic
  features.
\newblock \emph{NeurIPS}, 2019.

\bibitem[Eastwood et~al.(2021)Eastwood, Mason, Williams, and
  Sch{\"o}lkopf]{eastwood2021source}
Cian Eastwood, Ian Mason, Christopher~KI Williams, and Bernhard Sch{\"o}lkopf.
\newblock Source-free adaptation to measurement shift via bottom-up feature
  restoration.
\newblock \emph{Arxiv}, 2021.

\bibitem[Finn et~al.(2017)Finn, Abbeel, and Levine]{finn2017model}
Chelsea Finn, Pieter Abbeel, and Sergey Levine.
\newblock Model-agnostic meta-learning for fast adaptation of deep networks.
\newblock In \emph{ICML}, 2017.

\bibitem[Frikha et~al.()Frikha, Krompa{\ss}, and Tresp]{frikha2021arcade}
Ahmed Frikha, Denis Krompa{\ss}, and Volker Tresp.
\newblock Arcade: A rapid continual anomaly detector.
\newblock In \emph{2020 25th International Conference on Pattern Recognition
  (ICPR)}. IEEE.

\bibitem[Frikha et~al.(2021{\natexlab{a}})Frikha, Krompa{\ss}, K{\"o}pken, and
  Tresp]{frikha2021few}
Ahmed Frikha, Denis Krompa{\ss}, Hans-Georg K{\"o}pken, and Volker Tresp.
\newblock Few-shot one-class classification via meta-learning.
\newblock In \emph{Proceedings of the AAAI Conference on Artificial
  Intelligence}, 2021{\natexlab{a}}.

\bibitem[Frikha et~al.(2021{\natexlab{b}})Frikha, Krompa{\ss}, and
  Tresp]{frikha2021columbus}
Ahmed Frikha, Denis Krompa{\ss}, and Volker Tresp.
\newblock Columbus: Automated discovery of new multi-level features for domain
  generalization via knowledge corruption.
\newblock \emph{Arxiv}, 2021{\natexlab{b}}.

\bibitem[Ganin and Lempitsky(2015)]{ganin2015unsupervised}
Yaroslav Ganin and Victor Lempitsky.
\newblock Unsupervised domain adaptation by backpropagation.
\newblock In \emph{ICML}, 2015.

\bibitem[Ganin et~al.(2016)Ganin, Ustinova, Ajakan, Germain, Larochelle,
  Laviolette, Marchand, and Lempitsky]{ganin2016domain}
Yaroslav Ganin, Evgeniya Ustinova, Hana Ajakan, Pascal Germain, Hugo
  Larochelle, Fran{\c{c}}ois Laviolette, Mario Marchand, and Victor Lempitsky.
\newblock Domain-adversarial training of neural networks.
\newblock \emph{JMLR}, 2016.

\bibitem[Goodfellow et~al.(2014)Goodfellow, Shlens, and
  Szegedy]{goodfellow2014explaining}
Ian~J Goodfellow, Jonathon Shlens, and Christian Szegedy.
\newblock Explaining and harnessing adversarial examples.
\newblock \emph{Arxiv}, 2014.

\bibitem[Gretton et~al.(2012)Gretton, Borgwardt, Rasch, Sch{\"o}lkopf, and
  Smola]{gretton2012kernel}
Arthur Gretton, Karsten~M Borgwardt, Malte~J Rasch, Bernhard Sch{\"o}lkopf, and
  Alexander Smola.
\newblock A kernel two-sample test.
\newblock \emph{JMLR}, 2012.

\bibitem[Gulrajani and Lopez-Paz(2020)]{gulrajani2020search}
Ishaan Gulrajani and David Lopez-Paz.
\newblock In search of lost domain generalization.
\newblock \emph{Arxiv}, 2020.

\bibitem[He et~al.(2016)He, Zhang, Ren, and Sun]{he2016deep}
Kaiming He, Xiangyu Zhang, Shaoqing Ren, and Jian Sun.
\newblock Deep residual learning for image recognition.
\newblock In \emph{CVPR}, 2016.

\bibitem[Hinton et~al.(2015)Hinton, Vinyals, and Dean]{hinton2015distilling}
Geoffrey Hinton, Oriol Vinyals, and Jeff Dean.
\newblock Distilling the knowledge in a neural network.
\newblock \emph{Arxiv}, 2015.

\bibitem[Huang et~al.(2020)Huang, Wang, Xing, and Huang]{huang2020self}
Zeyi Huang, Haohan Wang, Eric~P Xing, and Dong Huang.
\newblock Self-challenging improves cross-domain generalization.
\newblock In \emph{ECCV}, 2020.

\bibitem[Hull(1994)]{hull1994database}
Jonathan~J. Hull.
\newblock A database for handwritten text recognition research.
\newblock \emph{IEEE Transactions on pattern analysis and machine
  intelligence}, 1994.

\bibitem[Ioffe and Szegedy(2015)]{ioffe2015batch}
Sergey Ioffe and Christian Szegedy.
\newblock Accelerating deep network training by reducing internal covariate
  shift.
\newblock In \emph{ICML}, 2015.

\bibitem[Kim et~al.(2021)Kim, Park, Kim, and Lee]{kim2021selfreg}
Daehee Kim, Seunghyun Park, Jinkyu Kim, and Jaekoo Lee.
\newblock Selfreg: Self-supervised contrastive regularization for domain
  generalization.
\newblock \emph{Arxiv}, 2021.

\bibitem[Kingma and Ba(2014)]{kingma2014adam}
Diederik~P Kingma and Jimmy Ba.
\newblock Adam: A method for stochastic optimization.
\newblock \emph{Arxiv}, 2014.

\bibitem[Kundu et~al.(2020)Kundu, Venkat, Babu, et~al.]{kundu2020universal}
Jogendra~Nath Kundu, Naveen Venkat, R~Venkatesh Babu, et~al.
\newblock Universal source-free domain adaptation.
\newblock In \emph{CVPR}, 2020.

\bibitem[LeCun et~al.(1998)LeCun, Bottou, Bengio, and
  Haffner]{lecun1998gradient}
Yann LeCun, L{\'e}on Bottou, Yoshua Bengio, and Patrick Haffner.
\newblock Gradient-based learning applied to document recognition.
\newblock \emph{IEEE}, 1998.

\bibitem[Li et~al.(2017)Li, Yang, Song, and Hospedales]{li2017deeper}
Da~Li, Yongxin Yang, Yi-Zhe Song, and Timothy~M Hospedales.
\newblock Deeper, broader and artier domain generalization.
\newblock In \emph{ICCV}, 2017.

\bibitem[Li et~al.(2018{\natexlab{a}})Li, Yang, Song, and
  Hospedales]{li2018learning}
Da~Li, Yongxin Yang, Yi-Zhe Song, and Timothy~M Hospedales.
\newblock Learning to generalize: Meta-learning for domain generalization.
\newblock In \emph{AAAI}, 2018{\natexlab{a}}.

\bibitem[Li et~al.(2018{\natexlab{b}})Li, Pan, Wang, and Kot]{li2018domain}
Haoliang Li, Sinno~Jialin Pan, Shiqi Wang, and Alex~C Kot.
\newblock Domain generalization with adversarial feature learning.
\newblock In \emph{CVPR}, 2018{\natexlab{b}}.

\bibitem[Li et~al.(2020)Li, Jiao, Cao, Wong, and Wu]{li2020model}
Rui Li, Qianfen Jiao, Wenming Cao, Hau-San Wong, and Si~Wu.
\newblock Model adaptation: Unsupervised domain adaptation without source data.
\newblock In \emph{CVPR}, 2020.

\bibitem[Li et~al.(2021{\natexlab{a}})Li, Jiang, Zhang, Kamp, and
  Dou]{li2021fedbn}
Xiaoxiao Li, Meirui Jiang, Xiaofei Zhang, Michael Kamp, and Qi~Dou.
\newblock Fedbn: Federated learning on non-iid features via local batch
  normalization.
\newblock \emph{arXiv preprint arXiv:2102.07623}, 2021{\natexlab{a}}.

\bibitem[Li et~al.(2018{\natexlab{c}})Li, Tian, Gong, Liu, Liu, Zhang, and
  Tao]{li2018deep}
Ya~Li, Xinmei Tian, Mingming Gong, Yajing Liu, Tongliang Liu, Kun Zhang, and
  Dacheng Tao.
\newblock Deep domain generalization via conditional invariant adversarial
  networks.
\newblock In \emph{ECCV}, 2018{\natexlab{c}}.

\bibitem[Li et~al.(2016)Li, Wang, Shi, Liu, and Hou]{li2016revisiting}
Yanghao Li, Naiyan Wang, Jianping Shi, Jiaying Liu, and Xiaodi Hou.
\newblock Revisiting batch normalization for practical domain adaptation.
\newblock \emph{Arxiv}, 2016.

\bibitem[Li et~al.(2021{\natexlab{b}})Li, Zhu, Gong, Shen, Dong, Yu, Lu, and
  Gu]{li2021mixmix}
Yuhang Li, Feng Zhu, Ruihao Gong, Mingzhu Shen, Xin Dong, Fengwei Yu, Shaoqing
  Lu, and Shi Gu.
\newblock Mixmix: All you need for data-free compression are feature and data
  mixing.
\newblock In \emph{ICCV}, 2021{\natexlab{b}}.

\bibitem[Liang et~al.(2020)Liang, Hu, and Feng]{liang2020we}
Jian Liang, Dapeng Hu, and Jiashi Feng.
\newblock Do we really need to access the source data? source hypothesis
  transfer for unsupervised domain adaptation.
\newblock In \emph{ICML}, 2020.

\bibitem[Liu et~al.(2021{\natexlab{a}})Liu, Chen, Qin, Dou, and
  Heng]{liu2021feddg}
Quande Liu, Cheng Chen, Jing Qin, Qi~Dou, and Pheng-Ann Heng.
\newblock Feddg: Federated domain generalization on medical image segmentation
  via episodic learning in continuous frequency space.
\newblock In \emph{Proceedings of the IEEE/CVF Conference on Computer Vision
  and Pattern Recognition}, pages 1013--1023, 2021{\natexlab{a}}.

\bibitem[Liu et~al.(2021{\natexlab{b}})Liu, Zhang, Wang, and Wang]{liu2021data}
Yuang Liu, Wei Zhang, Jun Wang, and Jianyong Wang.
\newblock Data-free knowledge transfer: A survey.
\newblock \emph{ArXiv}, 2021{\natexlab{b}}.

\bibitem[Lopes et~al.(2017)Lopes, Fenu, and Starner]{lopes2017data}
Raphael~Gontijo Lopes, Stefano Fenu, and Thad Starner.
\newblock Data-free knowledge distillation for deep neural networks.
\newblock \emph{Arxiv}, 2017.

\bibitem[Luo et~al.(2020)Luo, Sandler, Lin, Zhmoginov, and
  Howard]{luo2020large}
Liangchen Luo, Mark Sandler, Zi~Lin, Andrey Zhmoginov, and Andrew Howard.
\newblock Large-scale generative data-free distillation.
\newblock \emph{Arxiv}, 2020.

\bibitem[Mahendran and Vedaldi(2015)]{mahendran2015understanding}
Aravindh Mahendran and Andrea Vedaldi.
\newblock Understanding deep image representations by inverting them.
\newblock In \emph{CVPR}, 2015.

\bibitem[McInnes et~al.(2018)McInnes, Healy, and Melville]{mcinnes2018umap}
Leland McInnes, John Healy, and James Melville.
\newblock Umap: Uniform manifold approximation and projection for dimension
  reduction.
\newblock \emph{Arxiv}, 2018.

\bibitem[McMahan et~al.(2017)McMahan, Moore, Ramage, Hampson, and
  y~Arcas]{mcmahan2017communication}
Brendan McMahan, Eider Moore, Daniel Ramage, Seth Hampson, and Blaise~Aguera
  y~Arcas.
\newblock Communication-efficient learning of deep networks from decentralized
  data.
\newblock In \emph{Artificial intelligence and statistics}, pages 1273--1282.
  PMLR, 2017.

\bibitem[Micaelli and Storkey(2019)]{micaelli2019zero}
Paul Micaelli and Amos Storkey.
\newblock Zero-shot knowledge transfer via adversarial belief matching.
\newblock \emph{Arxiv}, 2019.

\bibitem[Mishra and Marr(2017)]{mishra2017apprentice}
Asit Mishra and Debbie Marr.
\newblock Apprentice: Using knowledge distillation techniques to improve
  low-precision network accuracy.
\newblock \emph{Arxiv}, 2017.

\bibitem[Motiian et~al.(2017)Motiian, Piccirilli, Adjeroh, and
  Doretto]{motiian2017unified}
Saeid Motiian, Marco Piccirilli, Donald~A Adjeroh, and Gianfranco Doretto.
\newblock Unified deep supervised domain adaptation and generalization.
\newblock In \emph{ICCV}, 2017.

\bibitem[Muandet et~al.(2013)Muandet, Balduzzi, and
  Sch{\"o}lkopf]{muandet2013domain}
Krikamol Muandet, David Balduzzi, and Bernhard Sch{\"o}lkopf.
\newblock Domain generalization via invariant feature representation.
\newblock In \emph{ICML}, 2013.

\bibitem[Nam et~al.(2019)Nam, Lee, Park, Yoon, and Yoo]{nam2019reducing}
Hyeonseob Nam, HyunJae Lee, Jongchan Park, Wonjun Yoon, and Donggeun Yoo.
\newblock Reducing domain gap via style-agnostic networks.
\newblock \emph{Arxiv}, 2019.

\bibitem[Nayak et~al.(2019)Nayak, Mopuri, Shaj, Radhakrishnan, and
  Chakraborty]{nayak2019zero}
Gaurav~Kumar Nayak, Konda~Reddy Mopuri, Vaisakh Shaj, Venkatesh~Babu
  Radhakrishnan, and Anirban Chakraborty.
\newblock Zero-shot knowledge distillation in deep networks.
\newblock In \emph{ICML}, 2019.

\bibitem[Netzer et~al.(2011)Netzer, Wang, Coates, Bissacco, Wu, and
  Ng]{netzer2011reading}
Yuval Netzer, Tao Wang, Adam Coates, Alessandro Bissacco, Bo~Wu, and Andrew~Y
  Ng.
\newblock Reading digits in natural images with unsupervised feature learning.
\newblock 2011.

\bibitem[Qiao et~al.(2020)Qiao, Zhao, and Peng]{qiao2020learning}
Fengchun Qiao, Long Zhao, and Xi~Peng.
\newblock Learning to learn single domain generalization.
\newblock In \emph{CVPR}, 2020.

\bibitem[Rajasegaran et~al.(2020)Rajasegaran, Khan, Hayat, Khan, and
  Shah]{rajasegaran2020self}
Jathushan Rajasegaran, Salman Khan, Munawar Hayat, Fahad~Shahbaz Khan, and
  Mubarak Shah.
\newblock Self-supervised knowledge distillation for few-shot learning.
\newblock \emph{Arxiv}, 2020.

\bibitem[Riemer et~al.(2018)Riemer, Cases, Ajemian, Liu, Rish, Tu, and
  Tesauro]{riemer2018learning}
Matthew Riemer, Ignacio Cases, Robert Ajemian, Miao Liu, Irina Rish, Yuhai Tu,
  and Gerald Tesauro.
\newblock Learning to learn without forgetting by maximizing transfer and
  minimizing interference.
\newblock \emph{ArXiv}, 2018.

\bibitem[Russakovsky et~al.(2015)Russakovsky, Deng, Su, Krause, Satheesh, Ma,
  Huang, Karpathy, Khosla, Bernstein, et~al.]{russakovsky2015imagenet}
Olga Russakovsky, Jia Deng, Hao Su, Jonathan Krause, Sanjeev Satheesh, Sean Ma,
  Zhiheng Huang, Andrej Karpathy, Aditya Khosla, Michael Bernstein, et~al.
\newblock Imagenet large scale visual recognition challenge.
\newblock \emph{International journal of computer vision}, 2015.

\bibitem[Shahtalebi et~al.(2021)Shahtalebi, Gagnon-Audet, Laleh, Faramarzi,
  Ahuja, and Rish]{shahtalebi2021sand}
Soroosh Shahtalebi, Jean-Christophe Gagnon-Audet, Touraj Laleh, Mojtaba
  Faramarzi, Kartik Ahuja, and Irina Rish.
\newblock Sand-mask: An enhanced gradient masking strategy for the discovery of
  invariances in domain generalization.
\newblock \emph{Arxiv}, 2021.

\bibitem[Shankar et~al.(2018)Shankar, Piratla, Chakrabarti, Chaudhuri, Jyothi,
  and Sarawagi]{shankar2018generalizing}
Shiv Shankar, Vihari Piratla, Soumen Chakrabarti, Siddhartha Chaudhuri, Preethi
  Jyothi, and Sunita Sarawagi.
\newblock Generalizing across domains via cross-gradient training.
\newblock \emph{Arxiv}, 2018.

\bibitem[Shi et~al.(2021)Shi, Seely, Torr, Siddharth, Hannun, Usunier, and
  Synnaeve]{shi2021gradient}
Yuge Shi, Jeffrey Seely, Philip~HS Torr, N~Siddharth, Awni Hannun, Nicolas
  Usunier, and Gabriel Synnaeve.
\newblock Gradient matching for domain generalization.
\newblock \emph{Arxiv}, 2021.

\bibitem[Sinha et~al.(2017)Sinha, Namkoong, Volpi, and
  Duchi]{sinha2017certifying}
Aman Sinha, Hongseok Namkoong, Riccardo Volpi, and John Duchi.
\newblock Certifying some distributional robustness with principled adversarial
  training.
\newblock \emph{Arxiv}, 2017.

\bibitem[Sun and Saenko(2016)]{sun2016deep}
Baochen Sun and Kate Saenko.
\newblock Deep coral: Correlation alignment for deep domain adaptation.
\newblock In \emph{ECCV}, 2016.

\bibitem[Torralba and Efros(2011)]{torralba2011unbiased}
Antonio Torralba and Alexei~A Efros.
\newblock Unbiased look at dataset bias.
\newblock In \emph{CVPR}, 2011.

\bibitem[Tzeng et~al.(2014)Tzeng, Hoffman, Zhang, Saenko, and
  Darrell]{tzeng2014deep}
Eric Tzeng, Judy Hoffman, Ning Zhang, Kate Saenko, and Trevor Darrell.
\newblock Deep domain confusion: Maximizing for domain invariance.
\newblock \emph{Arxiv}, 2014.

\bibitem[Wang et~al.(2020{\natexlab{a}})Wang, Shelhamer, Liu, Olshausen, and
  Darrell]{wang2020tent}
Dequan Wang, Evan Shelhamer, Shaoteng Liu, Bruno Olshausen, and Trevor Darrell.
\newblock Fully test-time adaptation by entropy minimization.
\newblock \emph{Arxiv}, 2020{\natexlab{a}}.

\bibitem[Wang et~al.(2020{\natexlab{b}})Wang, Li, and
  Kot]{wang2020heterogeneous}
Yufei Wang, Haoliang Li, and Alex~C Kot.
\newblock Heterogeneous domain generalization via domain mixup.
\newblock In \emph{ICASSP}, 2020{\natexlab{b}}.

\bibitem[Wilson and Cook(2020)]{wilson2020survey}
Garrett Wilson and Diane~J Cook.
\newblock A survey of unsupervised deep domain adaptation.
\newblock \emph{ACM TIST}, 2020.

\bibitem[Xu et~al.(2020)Xu, Zhang, Ni, Li, Wang, Tian, and
  Zhang]{xu2020adversarial}
Minghao Xu, Jian Zhang, Bingbing Ni, Teng Li, Chengjie Wang, Qi~Tian, and
  Wenjun Zhang.
\newblock Adversarial domain adaptation with domain mixup.
\newblock In \emph{AAAI}, 2020.

\bibitem[Yan et~al.(2020)Yan, Song, Li, Zou, and Ren]{yan2020improve}
Shen Yan, Huan Song, Nanxiang Li, Lincan Zou, and Liu Ren.
\newblock Improve unsupervised domain adaptation with mixup training.
\newblock \emph{Arxiv}, 2020.

\bibitem[Yeh et~al.(2021)Yeh, Yang, Yuen, and Harada]{yeh2021sofa}
Hao-Wei Yeh, Baoyao Yang, Pong~C Yuen, and Tatsuya Harada.
\newblock Source-data-free feature alignment for unsupervised domain
  adaptation.
\newblock In \emph{WACV}, 2021.

\bibitem[Yin et~al.(2020)Yin, Molchanov, Alvarez, Li, Mallya, Hoiem, Jha, and
  Kautz]{yin2020dreaming}
Hongxu Yin, Pavlo Molchanov, Jose~M Alvarez, Zhizhong Li, Arun Mallya, Derek
  Hoiem, Niraj~K Jha, and Jan Kautz.
\newblock Dreaming to distill: Data-free knowledge transfer via deepinversion.
\newblock In \emph{CVPR}, 2020.

\bibitem[Yin et~al.(2021)Yin, Zhu, and Hu]{yin2021comprehensive}
Xuefei Yin, Yanming Zhu, and Jiankun Hu.
\newblock A comprehensive survey of privacy-preserving federated learning: A
  taxonomy, review, and future directions.
\newblock \emph{ACM Computing Surveys (CSUR)}, 54\penalty0 (6):\penalty0 1--36,
  2021.

\bibitem[Zhang et~al.(2017{\natexlab{a}})Zhang, Cisse, Dauphin, and
  Lopez-Paz]{zhang2017mixup}
Hongyi Zhang, Moustapha Cisse, Yann~N Dauphin, and David Lopez-Paz.
\newblock mixup: Beyond empirical risk minimization.
\newblock \emph{Arxiv}, 2017{\natexlab{a}}.

\bibitem[Zhang et~al.(2021{\natexlab{a}})Zhang, Lei, Shi, Huang, and
  Chen]{zhang2021federated}
Liling Zhang, Xinyu Lei, Yichun Shi, Hongyu Huang, and Chao Chen.
\newblock Federated learning with domain generalization.
\newblock \emph{arXiv preprint arXiv:2111.10487}, 2021{\natexlab{a}}.

\bibitem[Zhang et~al.(2021{\natexlab{b}})Zhang, Qin, Ding, Gong, Yan, Tao, Li,
  Yu, and Liu]{zhang2021diversifying}
Xiangguo Zhang, Haotong Qin, Yifu Ding, Ruihao Gong, Qinghua Yan, Renshuai Tao,
  Yuhang Li, Fengwei Yu, and Xianglong Liu.
\newblock Diversifying sample generation for accurate data-free quantization.
\newblock In \emph{CVPR}, 2021{\natexlab{b}}.

\bibitem[Zhang et~al.(2017{\natexlab{b}})Zhang, David, and
  Gong]{zhang2017curriculum}
Yang Zhang, Philip David, and Boqing Gong.
\newblock Curriculum domain adaptation for semantic segmentation of urban
  scenes.
\newblock In \emph{ICCV}, 2017{\natexlab{b}}.

\bibitem[Zhao et~al.(2020)Zhao, Wang, Zhang, Gu, Li, Song, Xu, Hu, Chai, and
  Keutzer]{zhao2020multi}
Sicheng Zhao, Guangzhi Wang, Shanghang Zhang, Yang Gu, Yaxian Li, Zhichao Song,
  Pengfei Xu, Runbo Hu, Hua Chai, and Kurt Keutzer.
\newblock Multi-source distilling domain adaptation.
\newblock In \emph{AAAI}, 2020.

\bibitem[Zhou et~al.(2021{\natexlab{a}})Zhou, Liu, Qiao, Xiang, and
  Loy]{zhou2021domain}
Kaiyang Zhou, Ziwei Liu, Yu~Qiao, Tao Xiang, and Chen~Change Loy.
\newblock Domain generalization: A survey.
\newblock \emph{Arxiv}, 2021{\natexlab{a}}.

\bibitem[Zhou et~al.(2021{\natexlab{b}})Zhou, Yang, Qiao, and
  Xiang]{zhou2021mixstyle}
Kaiyang Zhou, Yongxin Yang, Yu~Qiao, and Tao Xiang.
\newblock Domain generalization with mixstyle.
\newblock \emph{Arxiv}, 2021{\natexlab{b}}.

\end{thebibliography}






\end{document}